\documentclass{article}

\usepackage{arxiv}

\usepackage[utf8]{inputenc} 
\usepackage[T1]{fontenc}    
\usepackage{hyperref}       
\usepackage{url}            
\usepackage{booktabs}       
\usepackage{amsfonts}       
\usepackage{nicefrac}       
\usepackage{microtype}      
\usepackage{amsmath}
\usepackage{natbib}
\usepackage{graphicx}
\title{Soft edit distance for differentiable comparison of symbolic sequences}

\author{
Evgenii Ofitserov \\
Department of Applied Mathematics and Computer Science\\
Tula State University\\
Lenina pr., 92, Tula 300012, Russia\\
\texttt{eofitserov@gmail.com}
\And
Vasily Tsvetkov \\
Pirogov Russian National Research Medical University (RNRMU) \\
Ostrovitianov str. 1, Moscow, Russia, 117997\\
Shemyakin-Ovchinnikov Institute of Bioorganic Chemistry of the Russian Academy of Sciences\\
Miklukho-Maklaya, 16/10, 117997, Moscow, Russian Federation \\
\And
Vadim Nazarov \\
National Research University Higher School of Economics, 20 Myasnitskaya Ulitsa, Moscow 101000, Russia\\
Keldysh Institute of Applied Mathematics, Miusskaya pl., 4, Moscow 125047, Russia\\}
\begin{document}
\maketitle

\begin{abstract}
Edit distance, also known as Levenshtein distance, is an essential way to compare two strings that proved to be particularly useful in the analysis of genetic sequences and natural language processing. However, edit distance is a discrete function that is known to be hard to optimize. This fact hampers the use of this metric in Machine Learning. Even as simple algorithm as K-means fails to cluster a set of sequences using edit distance if they are of variable length and abundance. In this paper we propose a novel metric --- soft edit distance (SED), which is a smooth approximation of edit distance. It is differentiable and therefore it is possible to optimize it with gradient methods. Similar to original edit distance, SED as well as its derivatives can be calculated with recurrent formulas at polynomial time. We prove usefulness of the proposed metric on synthetic datasets and clustering of biological sequences.
\end{abstract}


\section{Introduction}
	The problem of sequence comparison is fundamental in Machine Learning. It has arisen in many domains, including bioinformatics and Natural Language Processing (NLP), in such tasks as strings clustering, sequence to sequence learning and others. The most common approach to process similarity queries is the metric model due to its topological properties, although other models were proposed. \citep{gen} Sequence clustering, which is crucial for modern bioinformatics, especially in computational genomics, \citep {ref_cluster} requires the ability to compare strings of different lengths with respect to possible symbol insertions deletions and substitutions of symbols.
	
	Edit distance is an essential way to compare two strings. It is defined as minimal number of insertions, substitutions and deletions required to transform one string to another. This metric has numerous advantages that proved to be particularly useful in bioinformatics and natural language processing. \citep{ref_apr} Not only edit distance is intuitive and therefore highly interpretable, but it also can be efficiently computed with dynamic programming approach \citep{ref_WF}.
	
	However, edit distance is a discrete function that is known to be hard to optimize. This fact hampers the application of many Machine Learning methods to sequential data. Even as simple task as finding a centroid of a set of objects in sequence space with edit-distance metric presents significant difficulties. Whereas a centroid of n-vectors can be found merely using the arithmetic mean, a more sophisticated approach is needed to find a centroid or \textit{consensus} of a set with sequences of variable length $D$. For instance, it can be defined by following equation:
	\begin{equation}\label{centroid_task}
	c = \arg\min\limits_s\frac{1}{|D|}\sum\limits_{x\in D}d \left(x, s\right), \end{equation}
	where $d$ is edit distance. Task \eqref{centroid_task} is a discrete optimisation problem that becomes extremely complicated for large $D$. Existing methods of consensus search are based on multiple sequence alignment \citep{multiall} take a significant amount of computation time and cannot efficiently handle Big Data.
	
	A clustering problem with K-means algorithm is yet another example that illustrates the complexity of application of Machine Learning methods to sequential data. K-means is a simple and efficient method for clustering vector data, but it fails when input is a set of strings. As it has to recompute the centroids of said set on every iteration, the overall computation time grows exponentially. Similarly, any Machine Learning method that implies metric optimization is accompanied by discrete optimization problems, and thus it is not suitable for Big Data analysis in sequence space.
	
\section{Related work}	
    In this paper we propose a novel approach to sequence comparison based on soft edit distance. This metric is not subject to metric optimisation issues mentionned above. It can be optimised with continuous gradient methods, such as SGD or ADAM \citep{adam}, and it paves the way to the application of K-means and other Machine Learning algorithms to string data. 
    
    A string consists of discrete characters, and therefore we need to transform it into continuous set of objects in order to differentiate the metric for further comparison. To do this, we represent any sequence $x = x_{1}x_{2}\ldots x_{L}$, $x_{i} \in G$ by encoding matrix $X$ of shape $L \times |G|$, $X_{i,j} \in [0, 1]$, $\sum_j X_{i,j} = 1$. Each row of these matrices contains softmax parametrization of corresponding symbols, which corresponds to the one hot encoding for "hard defined" sequences. Such representation allows us to work with strings as with continuous objects.
    
\begin{figure}[h!]
        \begin{minipage}[t]{0.5\textwidth}
        \includegraphics[width=\textwidth]{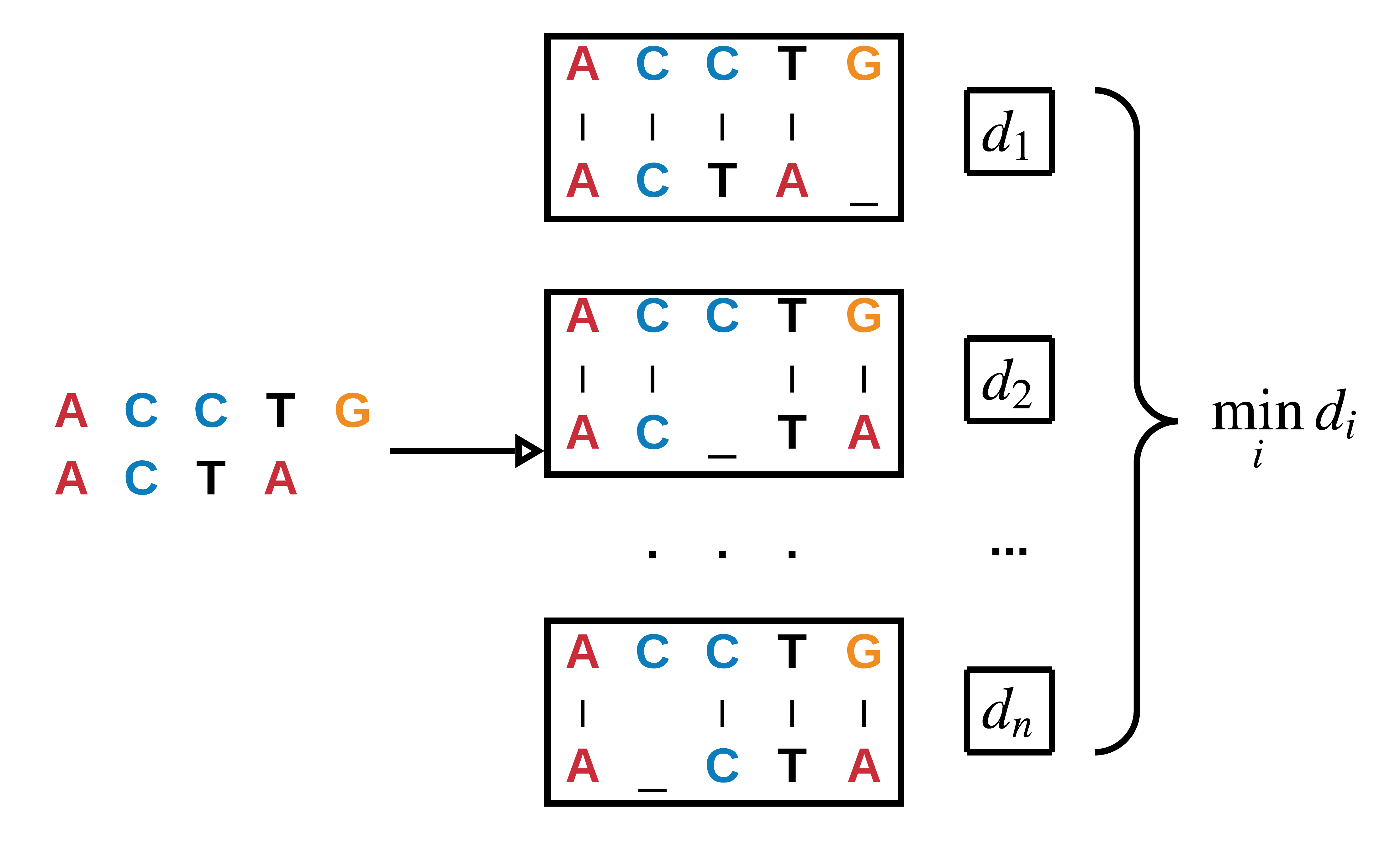}
        \end{minipage}
		\begin{minipage}[t]{0.5\textwidth}
        \includegraphics[width=\textwidth]{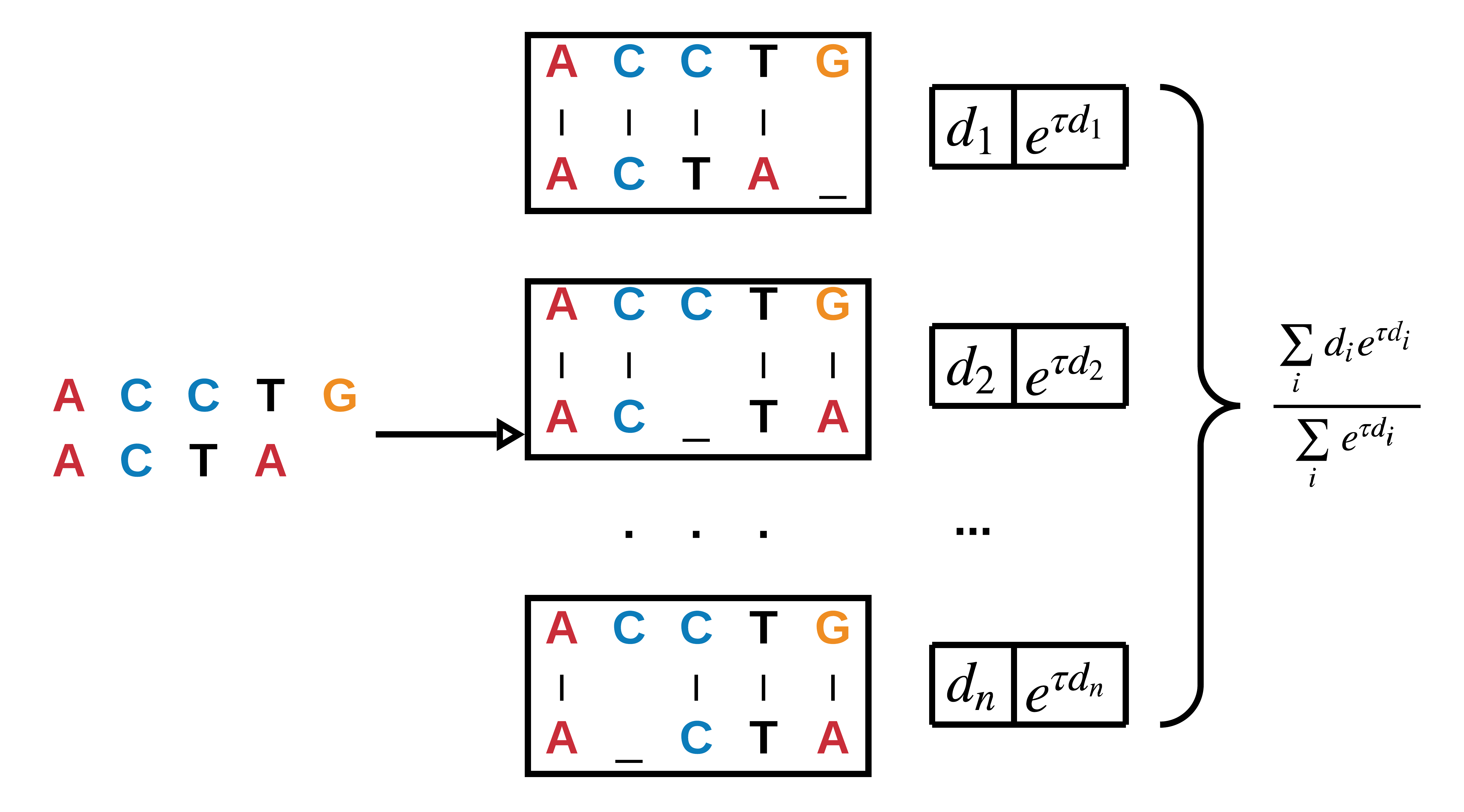}
        \end{minipage}
		\caption{Comparison of Levenshtein edit distance (left) and Soft edit distance (right). In contrast to standard edit distance SED is defined by softmin operation with parameter $\tau < 0$, which makes it subject to differentiation.} \label{fig:sed}
	\end{figure}

    After that, we introduce a novel metric --- soft edit distance (SED) that is a smooth approximation of original edit distance, which we are going to use to compare matrix representations of sequences. We define this metric by \textit{softmin} operation as weighted average of all possible ways to transform one string into another with insertions, substitutions and deletions Fig. \ref{fig:sed}. 
	By construction, proposed metric is smooth, and it can be differentiated with respect to the parameters of the encoding matrices. 
    
    In Section \ref{sec:rec} we propose recurrent formulas, which allow for the computation of SED and its derivatives at polynomial time similar to original edit distance.

	As we demonstrate further on, these concepts combined provide us with powerful means for sequential data clustering and consensus search. 
	
	\section{Soft edit distance}
	\subsection{Metric definition}
	Edit distance, also called Levenshtein distance, is an established method to compare strings. It can be formally defined as follows:
	
	Let $x_1 = x_{1,1}x_{1,2}\ldots x_{1,L_1}$, $x_2 = x_{2,1}x_{2,2}\ldots x_{2,L_2}$, $x_{j, i} \in G,~ L_j > 0$ --- two sequence in alphabet $G$. Edit distance between these two strings can be written in the form of following equation:
	\begin{equation}\label{edit_distance_def}
	ED(x_1, x_2)=\min\limits_{|x_1'|=|x_2'|=l} \left(d_H(x_1', x_2')+ L_1 - l + L_2 - l\right).
	\end{equation}
	Here we take minimum by all pair of subsequences of equal length  $x_1' \subset x_1$ and $x_2' \subset x_2$ with length $|x_1'|=|x_2'|$ respectively. Each of these pairs correspond to some alignment of sequences $x_1$ and $x_2$. $d_H(x_1', x_2')$ -- is a Hamming distance that indicates the number of mismatch positions in the substrings of equal length $x_1'$ and $x_2'$ and represents the number of substitutions in corresponding alignment. Similarly, components $L_1 - l$ and $L_2 - l$ indicate the number of insertions and deletions respectively. 
	
	The key idea of this paper is to replace all discrete operations in equation \eqref{edit_distance_def} with their smooth approximations and building soft approximation of edit distance. As we have already mentioned above, in order to build soft metric we will use matrix representation of input sequences $X_1$ and $X_2$ with shape $L_1\times|G|$ and $L_2\times|G|$ respectively. Hereafter we will use this representation for all sequences. In this notation $|X|$ means first dimension of matrix $X$, or length of corresponding sequence, and $X' \subset X$ indicate subset of rows of $X$ or matrix representation of corresponding subsequence $x'$.  
	
	Using such parametrization, we can replace Hamming distance $d_H(x_1', x_2')$ in equation \eqref{edit_distance_def} with element--wise absolute difference between corresponding matrices $X_1'$ and $X_2'$. With this in mind, the minimized value in the expression \eqref{edit_distance_def} can be written as:
	$$
	R(X_1', X_2') = \frac{1}{2}\sum\limits_{i=1}^{l} \sum\limits_{k=1}^{|G|} \left| X_{1,i,k}' - X_{2,i,k}' \right| + L_1 - l + L_2 - l,
	$$
	where $X_1' \subset X_1$ and $X_2' \subset X_2$ is a subsets of $l$ rows of original matrices $X_1$ and $X_2$. 
	
	The replacement of $\min$ operation in equation (\ref{edit_distance_def}) with \textit{softmin} function is a crucial step towards the derivation of soft edit distance formula. This gives us the following equation for soft edit distance:
	
	\begin{equation}\label{softedit_def}
	SED(X_1, X_2)=\frac{\sum\limits_{|X_1'|=|X_2'|} R(X_1', X_2') e^{\tau R(X_1', X_2')}}{\sum\limits_{|X_1'|=|X_2'|}  e^{\tau R(X_1', X_2')}},
	\end{equation} 
	where $\tau < 0$ --- softmin parameter. 
	
	The metric, defined by equation (\ref{softedit_def}), 
	meets the requirement of symmetry and non-negativity. However, in general $SED(X,X) \neq 0$. To overcome this problem we propose a non--biased soft edit distance that is defined by following rules:
	$$
	SED^0(X_1, X_2) = SED(X_1,X_2) - \frac{1}{2}\left(SED(X_1, X_1) + SED(X_2, X_2)\right).
	$$
	It's obvious that $SED^0(X, X) = 0~\forall X$.
	
	\subsection{Recurrence equation}\label{sec:rec}
	The na\"ive algorithm that computes $SED(X_1,X_2)$ by iterating through all pairs of subsequences of equal length has exponential-time complexity. To get around this issue we propose to use recursive function, which can be computed at polynomial time. As in Wagner--Fischer algorithm \citep{ref_WF} we iteratively compute distance between  all prefixes of the first string and all prefixes of the second by filling values of special matrices.
	
	We introduce the following notation. Let
	\begin{align*}
	&\alpha_{i,j}=\sum\limits_{\substack{|X_1'|=|X_2'| \\ X_1' \subset X_{1,1:i} \\ X_2' \subset X_{2,1:j}}} R_{i,j}(X_1', X_2') e^{\tau R(X_1', X_2')},
	\\
	&\beta_{i,j}=\sum\limits_{\substack{|X_1'|=|X_2'| \\ X_1' \subset X_{1,1:i} \\ X_2' \subset X_{2,1:j}}} e^{\tau R_{i,j}(X_1', X_2')},
	\\
	&i=\overline{0,L_1},~j=\overline{0,L_2},
	\end{align*}
	where $X_{1,1:i},~X_{2,1:j}$ are matrix representation of prefixes of sequences $x_1$ and $x_2$ with lengths $i$ and $j$ respectively. $L_1=|x_1|$, $L_2=|x_2|$ are lengths of $x_1$ and $x_2$. With this in mind, we will have:
	\begin{equation}\label{soft_edit_alphabeta}
	SED(X_1,X_2) = \frac{\alpha_{L_1, L_2}}{\beta_{L_1, L_2}}
	\end{equation}
	
	It can be shown, that the following recurrence equation is fulfilled for the coefficients $\alpha$ and $\beta$.  
	\begin{equation}\label{softedit_fw}
	\begin{aligned}
	&
	\begin{split}
	\alpha_{i,j}=\left( \alpha_{i-1,j} + \beta_{i-1,j} + \alpha_{i,j-1} + \beta_{i,j-1}\right )e^{\tau } + \left(\alpha_{i-1,j-1} +  \beta_{i-1,j-1} \delta_{i, j} \right)e^{\tau \delta_{i, j}} - \\ - \left(\alpha_{i-1,j-1} +  2\beta_{i-1,j-1} \right)e^{2\tau},~i=\overline{1, L_1},~j= \overline{1, L_2},
	\end{split}
	\\
	&\alpha_{i, 0} = ie^{\tau i},~i=\overline{0, L_1},
	\\
	&\alpha_{0, j} = je^{\tau j},~j=\overline{0, L_2},
	\\
	&\beta_{i,j} = \left( \beta_{i-1,j} + \beta_{i,j-1} \right)e^{\tau } + \beta_{i-1,j-1} \left(e^{\tau \delta_{i, j}} - e^{2\tau} \right), ~i=\overline{1, L_1},~j= \overline{1, L_2},
	\\
	&\beta_{i, 0} = e^{\tau i},~i=\overline{0, L_1},
	\\
	&\beta_{0, j} = e^{\tau j},~j=\overline{0, L_2}.
	\end{aligned}
	\end{equation}
	In the equation above $\delta_{i, j} = \frac{1}{2}\sum\limits_{k=1}^{|G|}|X_{1, i,k} - X_{2, j,k}|$. Complete inference of these formulas is given in Proofs section (\ref{app:theorem}). These recurrence formulas provide us with an efficient way to compute all values $\alpha_{i,j}$, $\beta_{i,j}$. Moreover, we can differentiate (\ref{softedit_fw}) with respect to $X_{\cdot,m,k}$ to get expression for $\frac{\partial \alpha_{i,j}}{\partial X_{\cdot,m,k}}$, $\frac{\partial \beta_{i,j}}{\partial X_{\cdot,m,k}}$, and compute derivative 
	$$
	\frac{\partial SED(x_1,x_2)}{\partial X_{\cdot,m,k}} = \frac{\frac{\partial \alpha_{L_1,L_2}}{\partial X_{\cdot,m,k}} \beta_{L_1,L_2} - \frac{\partial \beta_{L_1,L_2}}{\partial X_{\cdot,m,k}} \alpha_{L_1,L_2}}{\beta_{L_1,L_2}^2}.
	$$ 
	
	\section{Experiments}
	\subsection{Validation on synthetic dataset}
	In this section we validate SED on a synthetic dataset. We have created a synthetic dataset by randomly generating strings of length from 1 to 20 from an alphabet of DNA symbols $\{A, C, G, T\}$. Then for each pair of sequences we compare regular edit distance with a SED value. The results for different values of the softmin parameter $\tau$ are given in the Table \ref{tab1}. SED  converges to the regular edit distance metric with the decreasing of $\tau$, and for $\tau \leq 3$ the coefficient of determination $R^2 \geq 0.97$, which means extremely close approximation to the regular edit distance metric.  
	
	\begin{table}[h!]
		\caption{$R^2$ coefficient between classic edit distance and SED for different $\tau$ parameter, measured on set of $10^5$ random strings. All values are significant $(p < 0.01)$}\label{tab1}
		\begin{center}
			\begin{tabular}{l | l}
				$\tau$ &  $R^2$ coefficient\\
				\hline
				$\tau=-1$&  $0.49$ \\
				$\tau=-2$&  $0.93$ \\
				$\tau=-3$&  $0.97$ \\
				$\tau=-4$&  $0.99$ \\
				\hline
			\end{tabular}
		\end{center}
	\end{table}
	
	\subsection{Sequence clustering using K-means and SED}
	Our proposed smooth metric allows us to apply clustering algorithms such as K-means \citep{kmeans} to Big Data of symbolic sequences. In contrast to common string clustering algorithms, SED-based K-means makes it possible not only to break sequence sets into several clusters but to find directly the centroids or consensuses of those clusters.
	
	The key idea behind the SED-based K-means algorithm is to start with a fixed number of randomly initialized centroid sequences and then to perform two algorithmic steps in cycle. In the first step we randomly sample mini batch $M$ of sequences from the data set $D$ and mark them with labels corresponding to the nearest centroid. In a second step we use this mini batch and labels to update centroids. To achieve that we need to solve the following optimization problem:
	
	\begin{equation}\label{opt1}
	c^0_i=\arg\min\limits_{c}\sum\limits_{x \in M^i}SED^0(x, c),
	\end{equation}
	where $M^i \subset M$ is a set of strings from mini batch $M$, with label $i$, and $c^0_i$ is a matrix representation of corresponding centroid. Since SED is differentiable with the respect to the parameters of the input sequences, we can use stochastic gradient descent to solve (\ref{opt1}). 
	
	After clustering we transform matrix representation of centroid into symbolic consensus sequence by taking argmax in each row of matrix.
	
	In order to evaluate the proposed approach we tested SED-based K-means algorithm on different simulated sequences data sets. Firstly we explored the possibility of the proposed method to separate a mixture of sequences with different consensuses, depending of Levenshtein's distance between those consensuses.
	
	 For each experiment we generated a synthetic data set of strings of DNA symbols $\{A,C,G,T\}$ by the application of random noise --- insertions, substitutions and deletions to one of two arbitrary basis sequences $s_1$, $s_2$. After that, we separated generated sequences on a two clusters by our algorithm. 
	 
	 We conducted such tests for different lengths, noise rates and edit distances between $s_1$ and $s_2$ and measured accuracy of clustering and Levenstein's distance between true basis strings and inferred consensus of clusters. The results are given in Table \ref{tab2}. For each row of the table we conducted 10 experiments with different random basis sequences and averaged results.
	
	\begin{table}[h!]
		\caption{
		Result of dataset clustering. The dataset consisted of $20\cdot 10^3$ strings generated from two basis sequences. Basis length is a length of sequences that was used for data generation and $d(s_1, s_2)$ is edit distance between this two basis sequence; \textit{Noise rate} is a maximum number of insertions, deletions and substitutions in the resulting sequence; \textit{Acc}. is accuracy score of clustering; $\delta$ is a mean edit distance between inferred cluster consensuses and corresponding basis strings; $\epsilon$ is a percent of ideally matched consensuses (edit distance with corresponding basis string is equal to zero). Time is measured on Nvidia GPU GTX1080.}\label{tab2}
		\begin{center}
			\begin{tabular}{l | l | l | l | l | l | l}
				Basis length & $d(s_1, s_2)$ & Noise rate  & Acc. & $\delta$ & $\epsilon$ & Time\\
				\hline
				10 & 5 & 2 & 0.99 & 0.10 & 0.95 & 26.02 sec.\\
				10 & 5 & 5 & 0.99 & 0.00 & 1.00 & 27.72 sec.\\
				20 & 10 & 5 & 1.00 & 0.30 & 0.85 & 96.43 sec.\\
				20 & 10 & 10 & 0.99 & 0.80 & 0.60& 112.41 sec.\\
				30 & 15 & 7 & 1.00 & 1.20 & 0.40& 487.71 sec.\\
				30 & 15 & 15 & 1.00 & 1.10 & 0.50& 498.97 sec.\\
				\hline
			\end{tabular}
		\end{center}
	\end{table}
	
	Further, we conducted a series of experiments to evaluate the accuracy of clustering for bigger number of clusters. For this purpose we introduced more basis sequences. Results of clustering for 5, 10 and 30 centroids are given in the Table \ref{tab3}.
	
\begin{figure}[h!]
	   \begin{minipage}[t]{0.5\textwidth}
        \includegraphics[width=\textwidth]{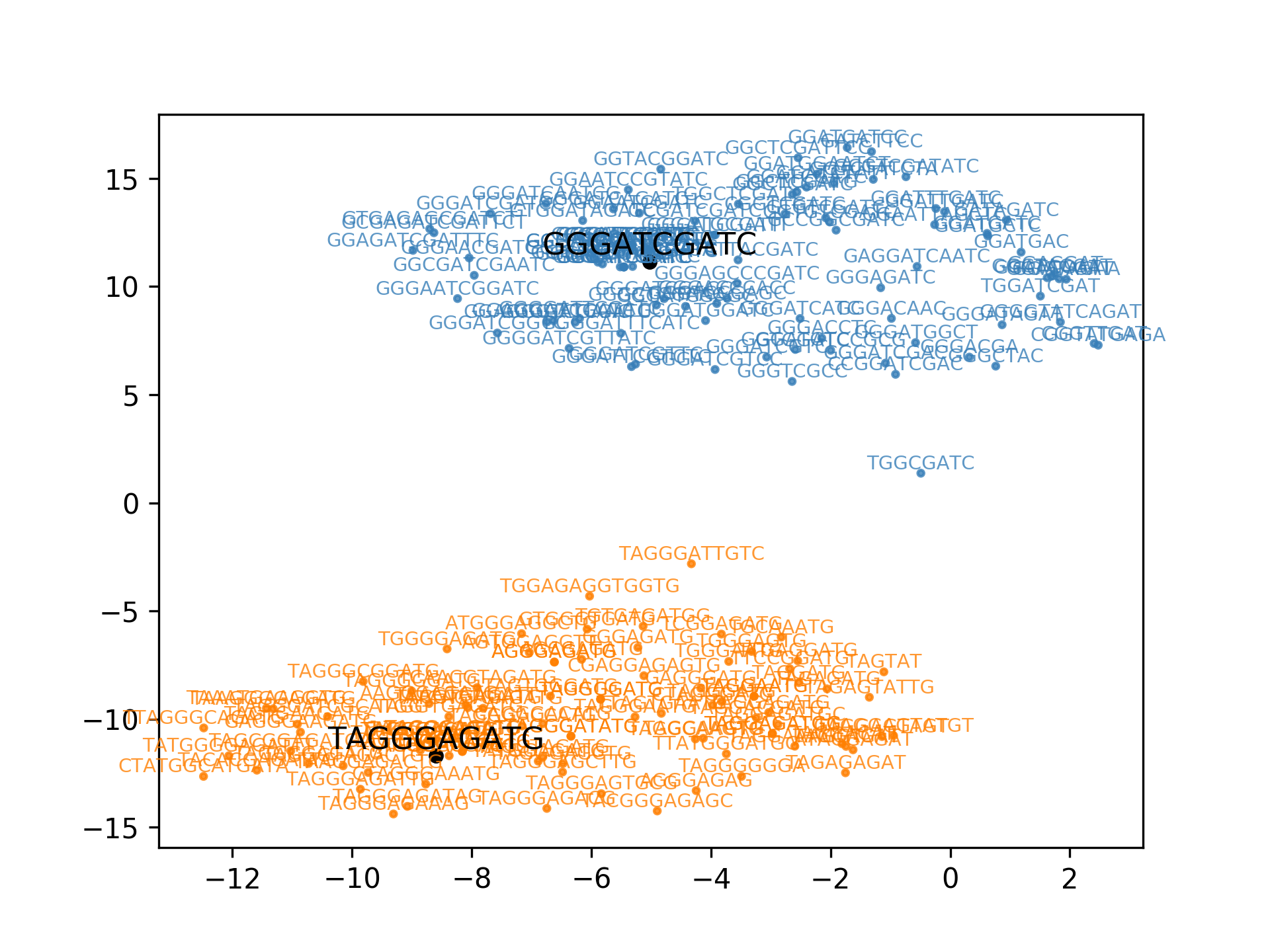}
        \end{minipage}
		\begin{minipage}[t]{0.5\textwidth}
        \includegraphics[width=\textwidth]{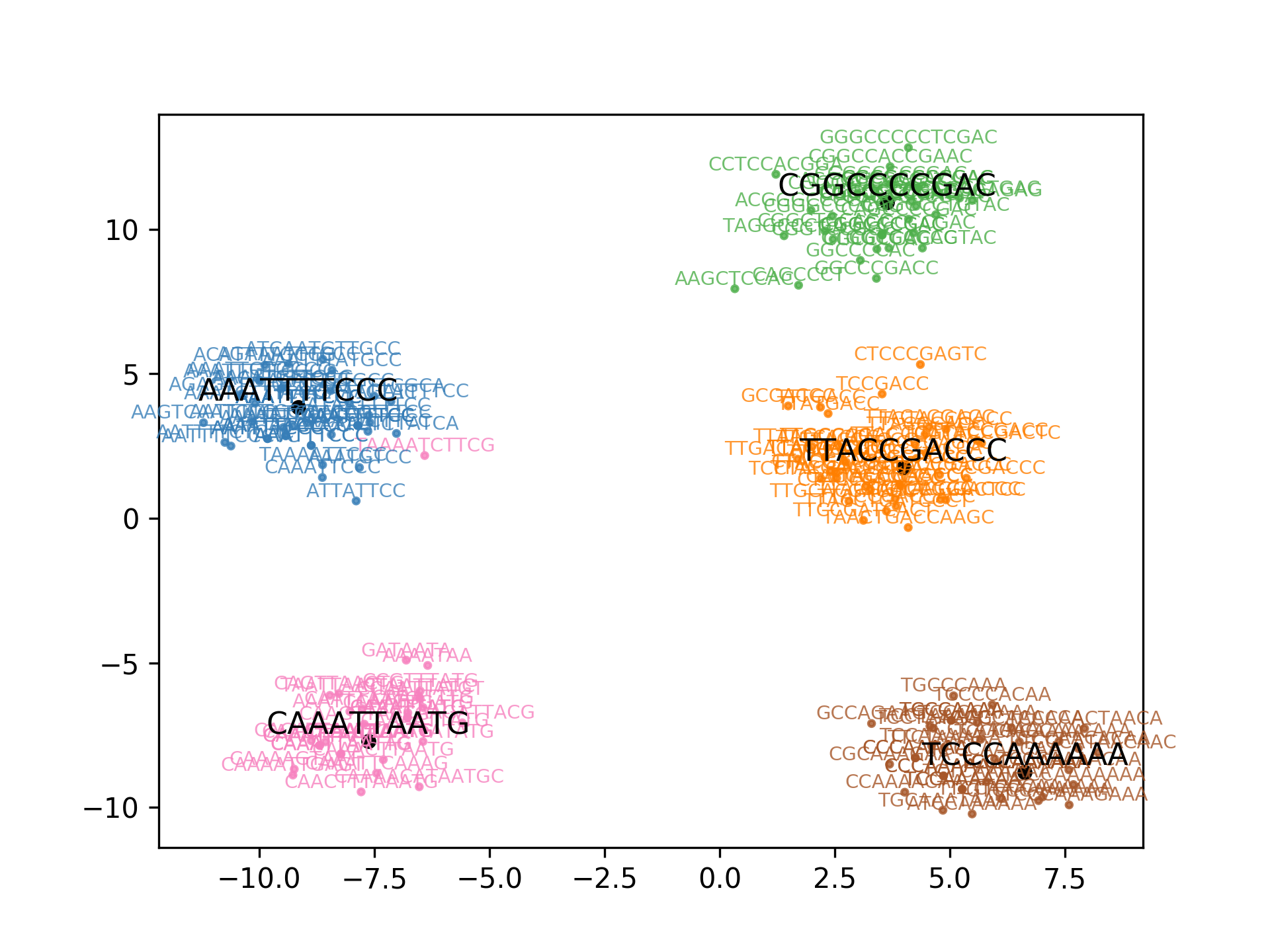}
        \end{minipage}
        \begin{minipage}[t]{0.5\textwidth}
        \includegraphics[width=\textwidth]{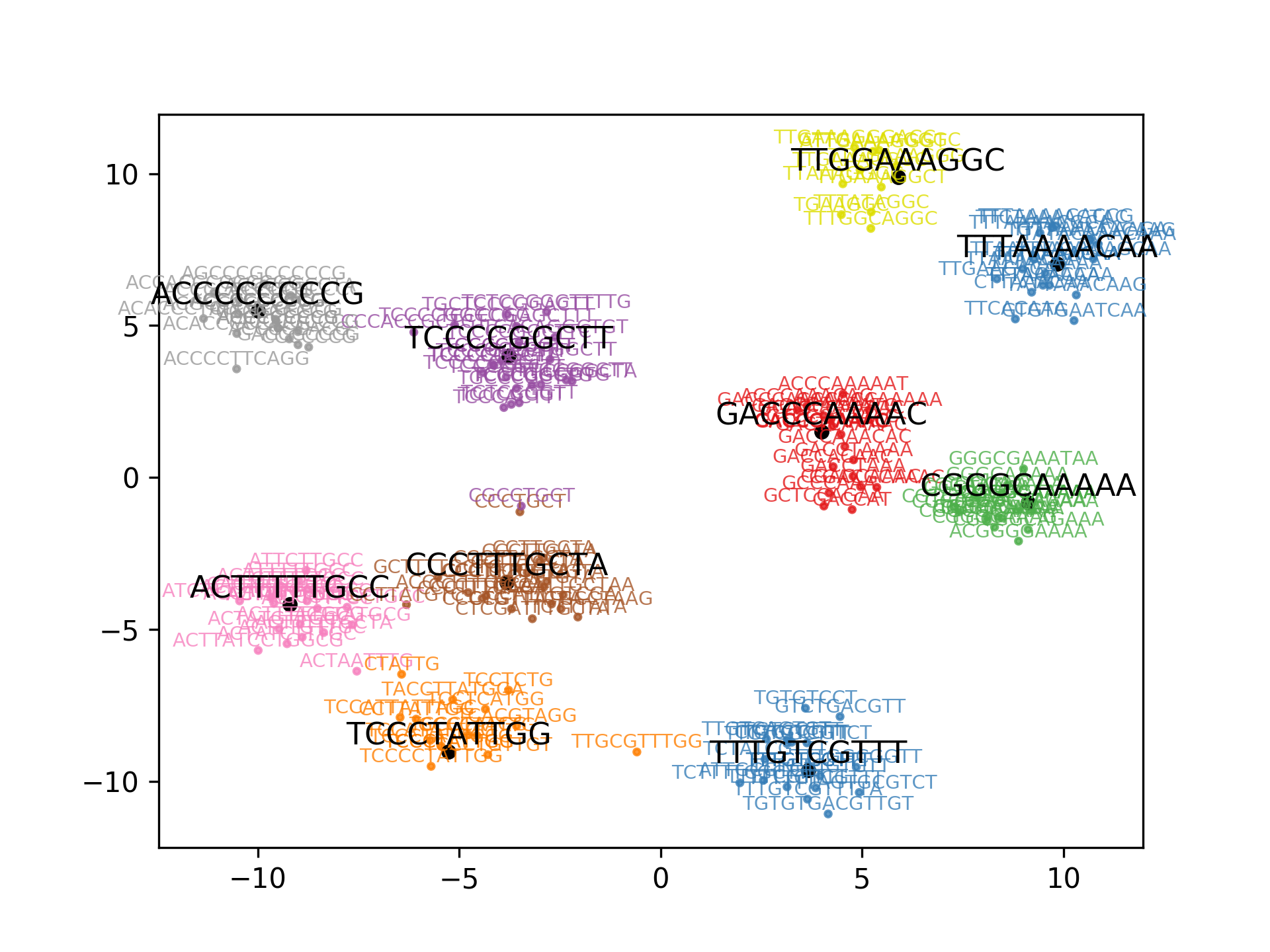}
        \end{minipage}

		\caption{Result of synthetic dataset clustering. Black points are centroids computed by our algorithm. Colours indicate assigned labels. We use the t-SNE projection to visualize centroids and 100 random strings from the original dataset.} \label{fig:simulated}
	\end{figure}

	Also, in Fig.\ref{fig:simulated} visualisations of tSNE\citep{tsne} projection of sub sample of clustered data sets and inferred consensuses are given. 
	
	\begin{table}[h!]
		\caption{Clustering results for different numbers of clusters. All experiments were conducted for basis sequences of length 10 and noise rate 5. Minimum edit distance between any two basis sequences is 5. \textit{N clusters} is a number of clusters; \textit{Acc}. is accuracy score of clustering; $\delta$ is a mean edit distance between inferred cluster consensuses and corresponding basis strings; $\epsilon$ is a percent of ideally matched consensuses (edit distance with corresponding basis string is equal to zero).}\label{tab3}
		\begin{center}
			\begin{tabular}{l | l | l | l}
				N clusters & Acc. & $\delta$ & $\epsilon$\\
				\hline
				5 & 0.971 & 0.240 & 0.920\\
				10 & 0.932 & 0.440 & 0.900\\
				30 & 0.971 & 0.707 & 0.840\\
				\hline
			\end{tabular}
		\end{center}
	\end{table}
	
\begin{figure}[h!]
		\includegraphics[width=\textwidth]{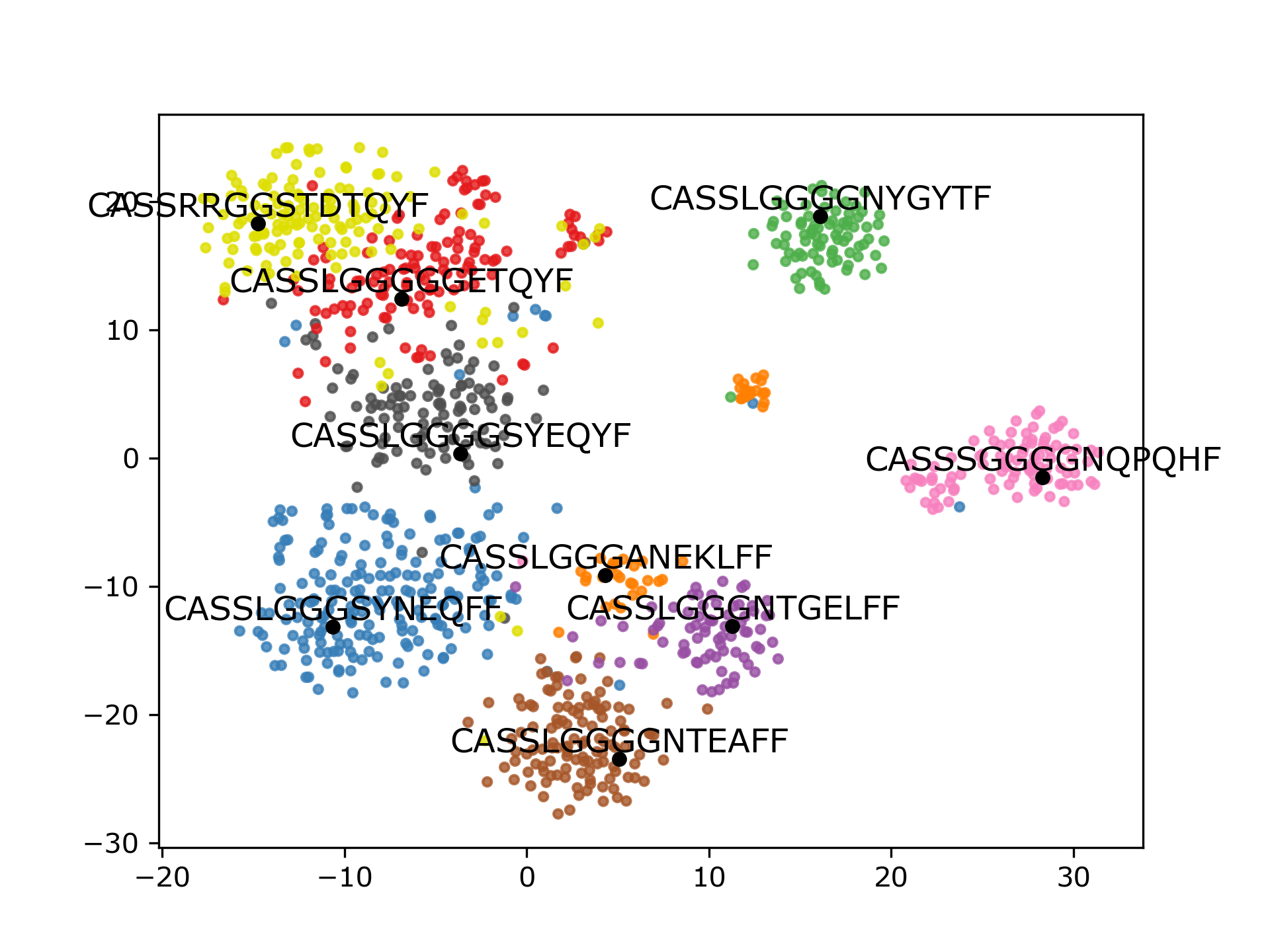}
		\caption{Visualisation of 200 clustered immune system sequences. Each coloured point is a string, assigned to one of the clusters. Black dots indicate cluster centroids as infered by the K-means algorithm.} \label{fig_repertoir}
	\end{figure}

	In order to test proposed method on a real data set, we have applied this approach to extract multiple consensus from set of immune system proteins called T-cell receptors. The dataset consists of $\approx 500\cdot10^3$ amino acids sequences --- strings of twenty-symbol alphabet with length varying from 10 to 20. The algorithm was used to cluster this data set on 9 clusters. Each cluster is represented by a centroid sequence of length 15. Visualisation of randomly selected sequences is given in Fig. \ref{fig_repertoir}.
	
	\section{Conclusion}
	In this paper, we have introduced a novel method for string comparison using differentiable edit distance. By virtue of that it is possible to optimize metric with stochastic gradient descent methods. As it was demonstrated, proposed soft edit distance can be used for fast consensus searching and  clustering of sequential Big Data with K-means algorithm.
	
	All described algorithms can be easily implemented and computed via GPU using existing deep learning frameworks. Source code of our Chainer \citep{ref:chainer} implementation can be found in \citep{ref:github}.
	
	In general, suggested metric can be used to solve the majority of problems that arise in computational biology regarding the consensus sequence discovery. Another possible application of proposed soft edit distance is Deep Learning. SED can be used ass loss function in sequence labeling task or in seq2seq learning as an alternative to the Connectionist Temporal Classification \citep{ref_CTC} approach. Also the SED can be applied to information retrieval from large databases and its representation. \citep{ref_ed_ext}

\newpage

\appendix
\section*{Appendix A.}
\label{app:theorem}

In this appendix we prove the following theorem from
Section~3.2:

\noindent
{\bf Theorem} {Coefficient $\alpha_{i,j},~\beta_{i,j}$ can be calculating by following recurrent rules:
		\begin{equation*}
		\begin{aligned}
		&
		\begin{split}
		\alpha_{i,j}=\left( \alpha_{i-1,j} + \beta_{i-1,j} + \alpha_{i,j-1} + \beta_{i,j-1}\right )e^{\tau } + \left(\alpha_{i-1,j-1} +  \beta_{i-1,j-1} \delta_{i, j} \right)e^{\tau \delta_{i, j}} - \\ - \left(\alpha_{i-1,j-1} +  2\beta_{i-1,j-1} \right)e^{2\tau},~i=\overline{1, L_1},~j= \overline{1, L_2}
		\end{split}
		\\
		&\alpha_{i, 0} = ie^{\tau i},~i=\overline{0, L_1}
		\\
		&\alpha_{0, j} = je^{\tau j},~j=\overline{0, L_2}
		\\
		&\beta_{i,j} = \left( \beta_{i-1,j} + \beta_{i,j-1} \right)e^{\tau } + \beta_{i-1,j-1} \left(e^{\tau \delta_{i, j}} - e^{2\tau} \right), ~i=\overline{1, L_1},~j= \overline{1, L_2}
		\\
		&\beta_{i, 0} = e^{\tau i},~i=\overline{0, L_1}
		\\
		&\beta_{0, j} = e^{\tau j},~j=\overline{0, L_2}
		\end{aligned}
		\end{equation*}
		, where $\delta_{i, j} = \frac{1}{2}\sum\limits_{k=1}^{|G|}|X_{1, i,k} - X_{2, j,k}|$.}

\noindent
{\bf Proof}. Let $\Omega_{1,i},~i=\overline{0, L_1}$ and $\Omega_{2,j},~j=\overline{0, L_2}$ --- are sets of matrix representations of all subsequences of prefixes $x_{1, 1:i}$ and $x_{2, 1:j}$ respectively. As we mentioned above, we represent all sequences in matrix form. Additionally, we introduce the following notation $X_{1, 1:i}$, $X_{2, 1:j}$ --- matrix representations of prefixes of sequences $x_1$ and $x_2$ with lengths $i$ and $j$ respectively. In this notation we will have: 
$$
\Omega_{1,i} = \{ X_1' ~|~ X_1' \subset X_{1, 1:i} \}
$$
$$
\Omega_{2,j} = \{ X_2' ~|~ X_2' \subset X_{2, 1:j} \}
$$

Then $\Omega_{1,i} \times \Omega_{2,j}$, is the Cartesian product of $\Omega_{1,i}$ and $\Omega_{2,j}$. Also, let introduce notation $E\left(\Omega_{1,i} \times \Omega_{2,j}\right) = \{(X_1', X_2')~|~(X_1',X_2') \in \Omega_{1,i} \times \Omega_{2,j},~|X_1'|=|X_2'|\}$ --- subset of $\Omega_{1,i} \times \Omega_{2,j}$ that include only pairs of equal length string.
		
		With this notation we will have: 
		\begin{align}\label{a_def}
		&\alpha_{i,j}=\sum\limits_{E(\Omega_{1,i} \times \Omega_{2,j})} R_{i,j}(X_1', X_2') e^{\tau R(X_1', X_2')}
		\\
		\label{b_def}
		&\beta_{i,j}=\sum\limits_{E(\Omega_{1,i} \times \Omega_{2,j})} e^{\tau R_{i,j}(X_1', X_2')}
		\\
		&R_{i, j}(X_1', X_2') = \sum\limits_{r=1}^{|X_1'|}\sum\limits_{s=1}^{|G|}|X_{1,r,s}'- X_{2,r,s}'| + |x_{1, 1:i}| - |X_1'| + |x_{2, 1:j}| - |X_2'|
		\end{align}
		
		Observing that $\Omega_{1,0}=\Omega_{2,0}=\{\emptyset \}$, where $\emptyset$ --- empty string, we can see that $\alpha_{i,0} = R_{i, 0}(\emptyset, \emptyset)e^{\tau} R_{i, 0}(\emptyset, \emptyset)$. From $\tilde{d}_H(\emptyset, \emptyset)=0$ follow that $\alpha_{i,0}=ie^{\tau i}$.
		
		In the same way (\ref{softedit_fw}) can be proofed for $\alpha_{0,j},~\beta_{i, 0},~\beta_{0,j}$.
		
		In case when $i>0$, $\Omega_{1,i}$ can be represented as $\Omega_{1,i} = \Omega_{1,i-1} \cup \Omega_{1,i}^*$, where $\Omega_{1,i}^*$ --- set of all subsequences of prefix $X_{1, 1:i}$, that contain $i$ symbol of sequence $x_1$. Similarly, $\Omega_{2,j} = \Omega_{2,j-1} \cup \Omega_{2,j}^*$ for $j>0$, than $\Omega_{1,i} \times \Omega_{2,j}$ can be represented as union of not intersected sets:
		\begin{equation*}
		\begin{split}
		\Omega_{1,i} \times \Omega_{2,j} =
		\left(\Omega_{1,i-1} \cup \Omega_{1,i}^*\right) \times \left(\Omega_{2,j-1} \cup \Omega_{2,j}^*\right)=\\=\left(\Omega_{1,i-1} \times \Omega_{2,j-1}\right) \cup \left(\Omega_{1,i-1} \times \Omega_{2,j}^*\right) \cup \left(\Omega_{1,i}^* \times \Omega_{2,j-1}\right) \cup \left(\Omega_{1,i}^* \times \Omega_{2,j}^*\right)
		\end{split}
		\end{equation*} 
		At the same time
		\begin{equation}\label{set_union}
		\begin{split}
		E\left(\Omega_{1,i} \times \Omega_{2,j}\right) =E\left(\Omega_{1,i-1} \times \Omega_{2,j-1}\right) \cup E\left(\Omega_{1,i-1} \times \Omega_{2,j}^*\right) \cup \\ \cup E\left(\Omega_{1,i}^* \times \Omega_{2,j-1}\right) \cup E\left(\Omega_{1,i}^* \times \Omega_{2,j}^*\right)
		\end{split}
		\end{equation} 
		
		So, as the sets in (\ref{set_union}) do not overlap, the relations (\ref{a_def}), (\ref{b_def}) can be rewritten in the form:
		\begin{equation}\label{alpha_set_sum}
		\begin{split}
		\alpha_{i,j}=\sum\limits_{E\left(\Omega_{1,i-1} \times \Omega_{2,j-1}\right)} A(X_1', X_2')  + \sum\limits_{ E\left(\Omega_{1,i-1} \times \Omega_{2,j}^*\right)}A(X_1', X_2') + \\ +
		\sum\limits_{ E\left(\Omega_{1,i}^* \times \Omega_{2,j-1}\right)} A(X_1', X_2') + \sum\limits_{ E\left(\Omega_{1,i}^* \times \Omega_{2,j}^*\right)} A(X_1', X_2'), 
		\end{split}
		\end{equation}
		where $A(X_1', X_2') = R_{i,j}(X_1', X_2') e^{\tau R_{i,j}(X_1', X_2')}$ and:
		\begin{equation}\label{beta_set_sum}
		\begin{split}
		\beta_{i,j}=\sum\limits_{E\left(\Omega_{1,i-1} \times \Omega_{2,j-1}\right)}B(X_1', X_2') + \sum\limits_{ E\left(\Omega_{1,i-1} \times \Omega_{2,j}^*\right)} B(X_1', X_2')  + \\ + 
		\sum\limits_{ E\left(\Omega_{1,i}^* \times \Omega_{2,j-1}\right)}B(X_1', X_2') + \sum\limits_{ E\left(\Omega_{1,i}^* \times \Omega_{2,j}^*\right)} B(X_1', X_2') 
		\end{split}
		\end{equation}
		where $B(X_1', X_2') = e^{\tau R_{i,j}(X_1', X_2')}$.
		
		All parts of equations (\ref{alpha_set_sum}) (\ref{beta_set_sum}) can be calculated by $\alpha_{i,j-1}$, $\alpha_{i-1,j}$, $\alpha_{i-1,j-1}$, $\beta_{i,j-1}$, $\beta_{i-1,j}$, $\beta_{i-1,j-1}$.
		From definition for $R_{i,j}(X_1', X_2')$ it follows that $R_{i,j}(X_1', X_2') = R_{i-1,j}(X_1', X_2') + 1 = R_{i,j-1}(X_1', X_2') + 1$, and 
		
		\begin{equation*}
		\begin{split}
		\sum\limits_{E\left(\Omega_{1,i-1} \times \Omega_{2,j-1}\right)} R_{i,j}(X_1', X_2') e^{\tau R_{i,j}(X_1', X_2')} = \sum\limits_{E\left(\Omega_{1,i-1} \times \Omega_{2,j-1}\right)} \left(R_{i-1,j-1}(X_1', X_2') + 2\right)\cdot\\\cdot e^{\tau R_{i,j}(X_1', X_2') + 2\tau} = \left(\alpha_{i-1,j-1} + 2\beta_{i-1,j-1}\right)e^{2\tau}.
		\end{split}
		\end{equation*}
		In the same way, we can show that:
		$$
		\sum\limits_{E\left(\Omega_{1,i-1} \times \Omega_{2,j-1}\right)} e^{\tau R_{i,j}(X_1', X_2')} = \sum\limits_{E\left(\Omega_{1,i-1} \times \Omega_{2,j-1}\right)} e^{\tau R_{i-1,j-1}(X_1', X_2') + 2\tau} = \beta_{i-1,j-1}e^{2\tau}.
		$$
		
		Any pairs of subsequence in set $ E\left(\Omega_{1,i}^* \times \Omega_{2,j}^*\right)$, can be expressed as  $$\left(\text{concat}\left(X_1',X_{1,i}\right),\text{concat}\left(X_2',X_{2,j}\right)\right)$$,
		where \textit{concat} --- operation of string concatenation, $X_{1,i}$ and $X_{2,j}$ rows of $X$ with index $i$ and $j$ respectively. $\left(X_1',X_2'\right) \in E\left(\Omega_{1,i-1} \times \Omega_{2,j-1}\right)$ --- pair of strings from set $ E\left(\Omega_{1,i-1} \times \Omega_{2,j-1}\right)$. For all sequence of this type: $$R_{i,j}\left(\text{concat}\left(X_1',X_{1,i}\right),\text{concat}\left(X_2',X_{2,j}\right)\right) = R_{i-1,j-1}\left(X_1',X_2'\right) + \delta_{i, j}.$$ It's mean that
		$$
		\sum\limits_{E\left(\Omega_{1,i}^* \times \Omega_{2,j}^* \right)} R_{i,j}(X_1', X_2') e^{\tau R_{i,j}(X_1', X_2')} = \left(\alpha_{i-1,j-1} + \delta_{i, j}\beta_{i-1,j-1}\right)e^{\tau \delta_{i, j}},
		$$
		$$
		\sum\limits_{E\left(\Omega_{1,i}^* \times \Omega_{2,j}^* \right)} e^{\tau R_{i,j}(X_1', X_2')} = \beta_{i-1,j-1}e^{\tau \delta_{i, j}},
		$$
		
		Set $\Omega_{1,i}^*$ can be represented as $\Omega_{1,i}^* = \Omega_{1,i} \setminus \Omega_{1,i-1}$, it's mean that  $$E\left(\Omega_{1,i}^* \times \Omega_{2,j-1}\right) = E\left(\Omega_{1,i} \times \Omega_{2,j-1}\right) \setminus E\left(\Omega_{1,i-1} \times \Omega_{2,j-1}\right).$$ Based on fact that $E\left(\Omega_{1,i-1} \times \Omega_{2,j-1}\right) \subset E\left(\Omega_{1,i} \times \Omega_{2,j-1}\right)$, summation by set $E\left(\Omega_{1,i}^* \times \Omega_{2,j-1}\right)$ can be writen as:
		\begin{equation*}
		\begin{split}
		\sum\limits_{ E\left(\Omega_{1,i}^* \times \Omega_{2,j-1}\right)} R_{i,j}(X_1', X_2') e^{\tau R_{i,j}(X_1', X_2')} = \sum\limits_{ E\left(\Omega_{1,i} \times \Omega_{2,j-1}\right)} R_{i,j}(X_1', X_2') e^{\tau R_{i,j}(X_1', X_2')} -\\- \sum\limits_{ E\left(\Omega_{1,i-1} \times \Omega_{2,j-1}\right)} R_{i,j}(X_1', X_2') e^{\tau R_{i,j}(X_1', X_2')} = \left(\alpha_{i,j-1} + \beta_{i,j-1}\right)e^\tau -\\- \left(\alpha_{i-1,j-1} + 2\beta_{i-1,j-1}\right)e^{2\tau}
		\end{split}
		\end{equation*}
		
		In the same way we can get following rules:
		\begin{align*}
		&\sum\limits_{ E\left(\Omega_{1,i}^* \times \Omega_{2,j-1}\right)} e^{\tau R_{i,j}(X_1', X_2')} = \beta_{i,j-1}e^\tau - \beta_{i-1,j-1}e^{2\tau}
		\\
		&\sum\limits_{ E\left(\Omega_{1,i-1} \times \Omega_{2,j}^*\right)} R_{i,j}(X_1', X_2') e^{\tau R_{i,j}(X_1', X_2')} = \left(\alpha_{i-1,j} + \beta_{i-1,j}\right)e^\tau - \left(\alpha_{i-1,j-1} + 2\beta_{i-1,j-1}\right)e^{2\tau}
		\\
		&\sum\limits_{ E\left(\Omega_{1,i-1} \times \Omega_{2,j}^*\right)} e^{\tau R_{i,j}(X_1', X_2')} = \beta_{i-1,j}e^\tau - \beta_{i-1,j-1}e^{2\tau}
		\end{align*}
		
		Finally, we can proof recurrent formulas (\ref{softedit_fw}) by substitution of above equations for components in (\ref{alpha_set_sum}) and (\ref{beta_set_sum}).

\vskip 0.2in
\bibliography{sample}

\begin{thebibliography}{12}
\providecommand{\natexlab}[1]{#1}
\providecommand{\url}[1]{\texttt{#1}}
\expandafter\ifx\csname urlstyle\endcsname\relax
  \providecommand{\doi}[1]{doi: #1}\else
  \providecommand{\doi}{doi: \begingroup \urlstyle{rm}\Url}\fi

\bibitem[Al~Aziz et~al.(2017)Al~Aziz, Alhadidi, and Mohammed]{ref_apr}
Md~Momin Al~Aziz, Dima Alhadidi, and Noman Mohammed.
\newblock Secure approximation of edit distance on genomic data.
\newblock \emph{BMC medical genomics}, 10\penalty0 (2):\penalty0 41, 2017.

\bibitem[Bernardes et~al.(2015)Bernardes, Vieira, Costa, and
  Zaverucha]{ref_cluster}
Juliana~S Bernardes, Fabio~RJ Vieira, Lygia~MM Costa, and Gerson Zaverucha.
\newblock Evaluation and improvements of clustering algorithms for detecting
  remote homologous protein families.
\newblock \emph{BMC bioinformatics}, 16\penalty0 (1):\penalty0 34, 2015.

\bibitem[Collingridge and Kelly(2012)]{multiall}
Peter~W Collingridge and Steven Kelly.
\newblock Mergealign: improving multiple sequence alignment performance by
  dynamic reconstruction of consensus multiple sequence alignments.
\newblock \emph{BMC bioinformatics}, 13\penalty0 (1):\penalty0 117, 2012.

\bibitem[Fuad(2012)]{gen}
Muhammad Marwan~Muhammad Fuad.
\newblock Towards normalizing the edit distance using a genetic
  algorithms--based scheme.
\newblock In \emph{International Conference on Advanced Data Mining and
  Applications}, pages 477--487. Springer, 2012.

\bibitem[Fuad and Marteau(2008)]{ref_ed_ext}
Muhammad Marwan~Muhammad Fuad and Pierre-Fran{\c{c}}ois Marteau.
\newblock The extended edit distance metric.
\newblock In \emph{2008 International Workshop on Content-Based Multimedia
  Indexing}, pages 242--248. IEEE, 2008.

\bibitem[Graves et~al.(2006)Graves, Fern{\'a}ndez, Gomez, and
  Schmidhuber]{ref_CTC}
Alex Graves, Santiago Fern{\'a}ndez, Faustino Gomez, and J{\"u}rgen
  Schmidhuber.
\newblock Connectionist temporal classification: labelling unsegmented sequence
  data with recurrent neural networks.
\newblock In \emph{Proceedings of the 23rd international conference on Machine
  learning}, pages 369--376. ACM, 2006.

\bibitem[Kingma and Ba(2014)]{adam}
Diederik~P Kingma and Jimmy Ba.
\newblock Adam: A method for stochastic optimization.
\newblock \emph{arXiv preprint arXiv:1412.6980}, 2014.

\bibitem[Maaten and Hinton(2008)]{tsne}
Laurens van~der Maaten and Geoffrey Hinton.
\newblock Visualizing data using t-sne.
\newblock \emph{Journal of machine learning research}, 9\penalty0
  (Nov):\penalty0 2579--2605, 2008.

\bibitem[Ofitserov(2019)]{ref:github}
Evgenii Ofitserov.
\newblock Soft edit distance, 2019.
\newblock URL \url{https://github.com/JenEskimos/soft_edit_distance}.

\bibitem[Sculley(2010)]{kmeans}
David Sculley.
\newblock Web-scale k-means clustering.
\newblock In \emph{Proceedings of the 19th international conference on World
  wide web}, pages 1177--1178. ACM, 2010.

\bibitem[Tokui et~al.(2015)Tokui, Oono, Hido, and Clayton]{ref:chainer}
Seiya Tokui, Kenta Oono, Shohei Hido, and Justin Clayton.
\newblock Chainer: a next-generation open source framework for deep learning.
\newblock In \emph{Proceedings of workshop on machine learning systems
  (LearningSys) in the twenty-ninth annual conference on neural information
  processing systems (NIPS)}, volume~5, pages 1--6, 2015.

\bibitem[Wagner and Fischer(1974)]{ref_WF}
Robert~A Wagner and Michael~J Fischer.
\newblock The string-to-string correction problem.
\newblock \emph{Journal of the ACM (JACM)}, 21\penalty0 (1):\penalty0 168--173,
  1974.

\end{thebibliography}

\end{document}